\documentclass{article}
\usepackage{ijcai23}

\usepackage{times}
\usepackage{soul}
\usepackage{url}
\usepackage[hidelinks]{hyperref}
\usepackage[utf8]{inputenc}
\usepackage[small]{caption}
\usepackage{graphicx}
\usepackage{amsmath}
\usepackage{amsthm}
\usepackage{amsfonts} 
\usepackage{booktabs}
\usepackage{algorithm}
\usepackage{algorithmic}
\usepackage[switch]{lineno}
\usepackage{multicol}
\usepackage{multirow}
\usepackage{verbatim}

\newtheorem{theorem}{Theorem}
\newtheorem{definition}[theorem]{Definition}

\title{Long-term Wind Power Forecasting with Hierarchical Spatial-Temporal Transformer}

\author{
Yang Zhang$^1$
\and
Lingbo Liu$^{2,3}$\thanks{Corresponding author: Lingbo Liu (lingbo.liu@polyu.edu.hk).}\and
Xinyu Xiong$^1$\and
Guanbin Li$^{1,4}$ \and
Guoli Wang$^1$ \And
Liang Lin$^1$
\affiliations
$^1$School of Computer Science and Engineering, Sun Yat-sen University, Guangzhou, China\\
$^2$Department Land Surveying and Geo-Informatics, The Hong Kong Polytechnic University, Hong Kong\\
$^3$Smart Cities Research Institute, The Hong Kong Polytechnic University, Hong Kong\\
$^4$Research Institute, Sun Yat-sen University, Shenzhen, China
}
\begin{document}

\maketitle

\begin{abstract}
    Wind power is attracting increasing attention around the world due to its renewable, pollution-free, and other advantages. However, safely and stably integrating the high permeability intermittent power energy into electric power systems remains challenging. Accurate wind power forecasting (WPF) can effectively reduce power fluctuations in power system operations. Existing methods are mainly designed for short-term predictions and lack effective spatial-temporal feature augmentation. In this work, we propose a novel end-to-end wind power forecasting model named \underline{H}ierarchical \underline{S}patial-\underline{T}emporal \underline{T}ransformer \underline{N}etwork (HSTTN) to address the long-term WPF problems. Specifically, we construct an hourglass-shaped encoder-decoder framework with skip-connections to jointly model representations aggregated in hierarchical temporal scales, which benefits long-term forecasting. Based on this framework, we capture the inter-scale long-range temporal dependencies and global spatial correlations with two parallel Transformer skeletons and strengthen the intra-scale connections with downsampling and upsampling operations. Moreover, the complementary information from spatial and temporal features is fused and propagated in each other via Contextual Fusion Blocks (CFBs) to promote the prediction further. Extensive experimental results on two large-scale real-world datasets demonstrate the superior performance of our HSTTN over existing solutions.
\end{abstract}

\section{Introduction}
The UN Sustainable Development Goals 7 ~\cite{vinuesa2020role} (SDG 7) aims to ensure access to affordable, reliable, sustainable, and modern energy for all. Wind Power Forecasting (WPF), which focuses on accurately predicting the wind power generation of turbines in a wind farm for future time intervals, can contribute to the realization of SGD 7 by building more efficient low-carbon systems. It was reported that in 2021, the proportion of global wind and solar power generation had reached one-tenth, and the total energy output exceeds 2837 TWh \cite{Emberglobal}.However, due to the chaotic nature of the earth’s atmosphere, wind power generation is always associated with non-stationary uncertainties. Therefore, how to integrate wind energy into the power grid with high stability and security is of great significance.

Fortunately, these uncertainties in power systems operations can be mitigated to a certain degree via accurate WPF methods, which are becoming the most promising solutions for integrating a large amount of wind energy into power grids~\cite{Tastu}. Wind Power Forecasting has been extensively investigated over the past decades~\cite{DengShao,WangZou}, and the existing research can be coarsely divided into four categories: physics-based methods \cite{ChenQian,ShaoDeng}, statistical methods~\cite{ZengQiao,HuZhang}, hybrid intelligent methods~\cite{Ghoushchi,de2021daft}, and deep learning-based methods~\cite{ahmad2022data,ZhuChen}. However, most of the existing works still suffer from several limitations, which restricts their applications in real world:
\begin{itemize}
    \item Physics-based and statistical methods usually take too much calculation costs and are sensitive to the errors introduced by the initial condition~\cite{DengShao}. They can not perform well when dealing with nonlinear and non-stationary traits in wind power due to their shallow learning models~\cite{WangLi}.
    \item Most of them are designed for short-term predictions and can not achieve satisfactory results under long-term wind power forecasting~\cite{ShaoDeng,HeWang}, while accurate long-term predictions are even more critical in system dispatch planning and ramp events (large wind power fluctuation) prediction~\cite{OuyangHuang}.
    \item Existing methods lack an effective design for spatiotemporal modeling. They either consider WPF as a simple times series forecasting problem ignoring spatial information or extract spatial dependencies in a local and static manner~\cite{yu2020probabilistic,ZhuChen}. Actually, indispensable information may be revealed by modeling spatial correlations because wind characteristics at a site resemble those nearby or share the same meteorological conditions~\cite{DingCRC}. It is nontrivial to capture global and comprehensive spatial correlations.
\end{itemize}

The long-term WPF aims to understand the correlation between data in each different time step. However, 1) a single point does not have semantic meaning like a word in a sentence and may have limited influence on predicting the future~\cite{nie2022time}. What is more, 2) the inherent high intermittent of wind power and the record noises caused by some external reasons bring in plenty of uncertainties, which makes the long-term prediction worse. 3) the fine-grained long-term prediction leads to high computational and space complexities when applying the point-wise self-attention mechanism. In contrast, sparse and localized contextual information is essential in analyzing their semantic connections~\cite{du2022preformer}, e.g., the airflow in a period, like midnight, may show intense fluctuations but blows much heavier than that at noon. Thereby, the aggregated and coarse-grained temporal representations are nontrivial and complementary to the fine-grained temporal features for precise forecasting.

Inspired by the ideas and problems mentioned above, we propose a novel end-to-end deep learning-based framework termed Hierarchical Spatial-Temporal Transformer Network (HSTTN), which well addressed the long-term predictions with the hourglass-shaped network and effectively modeled the spatiotemporal contextual information and correlations. In particular, the HSTTN consists of four main modules: hourglass-shaped encoder-decoder architecture, residual spatiotemporal encoder/decoder layers, and Contextual Fusion Blocks (CFBs).

Different from the standard transformer architecture, in encoder, temporal pooling operations are inserted between several cascaded residual spatiotemporal encoder layers to generate hierarchical temporal scale features from fine-grain to coarse-grain. Symmetrically, in decoder, we gradually recover the fine-grained predictions from coarse-grained representations with upsampling operations inserted between the residual spatiotemporal decoder layers. Aggregating time steps to coarse-grained scale not only provides comprehensive semantic representations that are complementary to finer scales, but also reduces the amount of calculation since the sequence length is smaller. Besides, via inter-scale skip-connections, the outputs of each residual spatiotemporal encoder layer are directly concatenated to the outputs of each residual spatiotemporal decoder layer with the same temporal scales, which aggregates the rich fine-grained information from hierarchical encoder layers to facilitate the decoder to make more precise predictions.
It is noteworthy that vanilla Transformer ~\cite{VaswaniSha} is designed for the machine translation task which follows a seq2seq paradigm, while wind power records are spatiotemporal structured. So we first decouple the topological data into temporal-wise feature vectors and spatial-wise feature vectors, then feed them into the residual spatiotemporal encoder layers in parallel, which consists of temporal and spatial Transformer skeletons and contextual fusion blocks, to capture the hierarchical long-range temporal dependencies and spatial global correlations with the multi-head self-attention mechanism.
The CFBs are designed for better spatiotemporal feature fusion and are inserted between each temporal encoder layer and spatial encoder layer. In specific, the latent representations from both layers are firstly rearranged into original spatiotemporal-shape and concatenated along feature dimension. Then a Convolutional Neural Network is set to fuse and learn their contextual feature representations. The enhanced features which carry more comprehensive information are then fed into the next residual encoder layer for sparser scale but higher-level representation learning. The residual spatiotemporal decoder layers follow the same structure while the inputs are only future time spots and turbines locations without knowing the meteorological data and turbine internal status. 
The main contributions of this work are as follows:
\begin{itemize}
    \item We propose a Transformer-based framework to predict wind power generations, which well addresses the long-term forecasting problem due to its impactful capacity to capture long-range and global dependencies by self-attention mechanism. To the best of our knowledge, this is the first attempt to apply Transformer architecture to long-term wind power forecasting tasks.
    \item A well-designed hourglass-shaped hierarchical architecture with lower time and space complexity is introduced to improve the long-term predictions by aggregating complementary multiscale temporal representations.
    \item Extensive experiments are conducted on two real-world wind power datasets with diverse dynamic context factors (meteorological data and turbine internal status), which demonstrate the effectiveness of the proposed framework for wind power forecasting.

\end{itemize}

\section{Related Work}\label{sec:related_work}
\textbf{Wind Power Forecasting: }
Existing forecasting models can be grouped into four types based on differences in modeling theory: physics-based methods \cite{LoboSanchez,ShaoDeng}, statistical methods~\cite{ZengQiao,HuZhang}, hybrid intelligent methods \cite{HeWang,ShahidZam} and deep learning based methods \cite{DengShao,WangZou,WangLi}. In physical models, numerical weather predictions (NWP) or weather researcher forecasting (WRF) are usually performed to predict weather conditions and then the weather condition predictions are fed into physics-based models to generate wind power forecasting. Similarly, a statistical method was proposed by \cite{ZengQiao}, which first predicted the wind speed with a SVM-based model, then used the power-wind speed characteristic of the wind turbine generators to predict the final short-term wind power. A non-negligible drawback of these two-stage prediction frameworks is that the errors in the first stage will accumulate and magnify the final prediction errors. To avoid the limitations of a single model, hybrid intelligent method also attracted much attention, which is a weighted sum of several models or the combination of compensatory models. \cite{ShahidZam} developed a hybrid framework comprising of long short term memory (LSTM) and genetic algorithm (GA). The global optimization strategy of GA was exploited to optimize hyperparameters in LSTM layers. The deep learning based methods have drawn increasing attention in recent years due to its capacity of modeling intricate and non-linear relations. For instance, \cite{yu2020probabilistic} proposed a hybrid neural network to capture spatial-temporal characteristics, in which the spatial features were extracted by a 2D-CNN and the temporal features were extracted by an LSTM. CNN is efficient in local feature extraction whereas in this work we prefer global spatial correlations as illustrated in the above section. \cite{HongRio} proposed a convolutional neural network (CNN)-based model to extract wind power characteristics and a radial basis function neural network  with a double Gaussian function as its activation function was used to deal with uncertain characteristics.

\begin{figure*}
    \centering
    \includegraphics[width=1.00\linewidth]{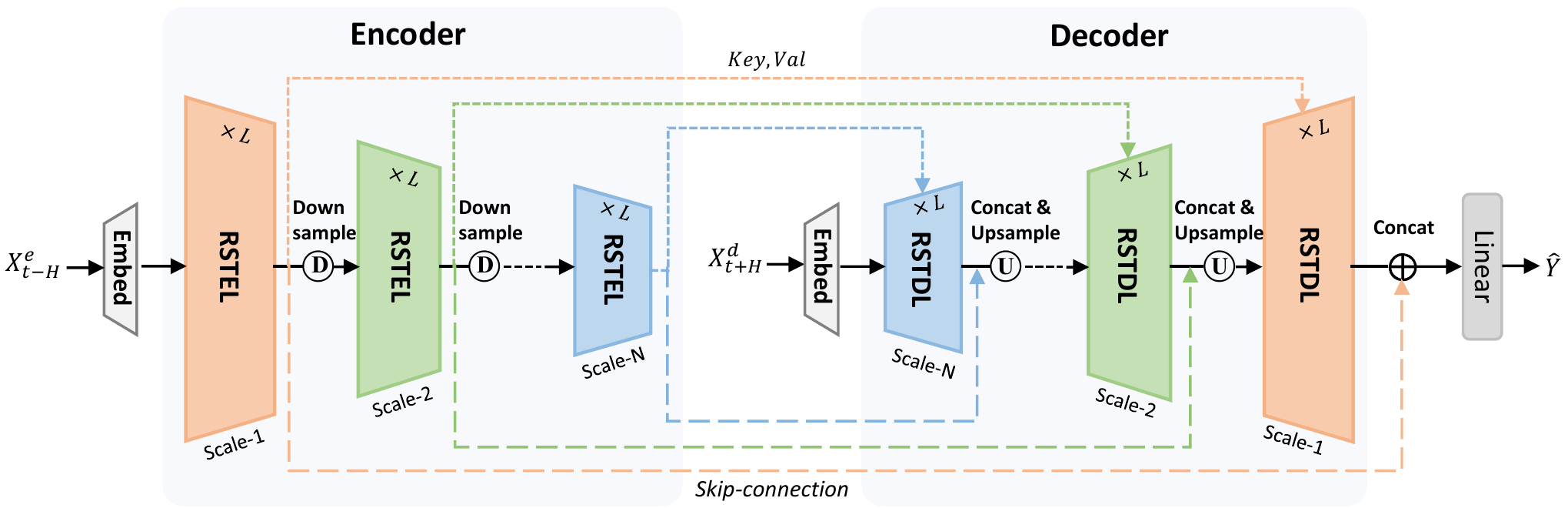}
    \vspace{-7mm}
    \caption{
    The architecture of our proposed Hierarchical Spatial-Temporal Transformer Network (HSTTN). RSTEL/RSTDL represent the residual spatiotemoral encoder/decoder layers, respectively. $\times L$ means each encoder/decoder layer could repeat multiple times for deeper semantics. $Key$ and $Val$ indicate the outputs of the encoder layers are mapped to latent spaces and transferred to their inter-scale decoder layers for attention calculations.
    }
    \vspace{-3mm}
    \label{fig:overview}
\end{figure*}

\noindent\textbf{Transformers: }
Transformer is an advanced attention-based neural network block, which is originally proposed to tackle the machine translation task then widely used in natural language processing due to its superior performance in capturing long-range dependency by global self-attention mechanism \cite{VaswaniSha,Bert}. Recently, researchers have applied transformers to more artificial intelligence tasks such as visual understanding \cite{li2021groupformer,wu2022multimodal,liu2022umt,li2022view,Zhang_2023_CVPR,jiang2023clip}, time series analysis \cite{ZhouZhang,wu2021autoformer,jiang2023transformer,luo2023end} and spatial-temporal modeling \cite{xu2020spatial,liu2021road,liu2022online,geng2022rstt}. For long-term forecasting, \cite{ZhouZhang} proposed a transformer-based model named Informer to predict long sequences and designed the ProbSparse self-attention mechanism and distilling operation which drastically improved the inference speed of long-sequence predictions. Others explore to divide time steps into segments and capture segment-wise correlations \cite{nie2022time,du2022preformer}. Inspired by these amazing works, we propose a Transformer-based spatial-temporal learning framework to hierarchically capture the contextual information interaction between complementary spatial and temporal features. To the best of our knowledge, this work is the first attempt to apply Transformer architecture to long-term wind power forecasting.

\section{Problem Specification}\label{sec:problem_specification}
In this section, we provide basic notations and the definition of wind power forecasting. The objective of WPF is to accurately estimate the wind power supply of a wind farm at different time steps by characterising the intricate relations between historical records and future wind power generations. In practice, wind is deflected by the blades of a wind turbine, and then generates electricity through rotations and generators, indicating the wind power is not only related to wind speed, but also to other meteorological data like wind direction, external temperature and essential turbine internal status.
Given a wind farm consisting of $N$ turbines, each of them generates wind power time series and corresponding dynamic context factors (meteorological data, turbine internal status, etc.). The dynamic context records of all turbines in a time window with $T$ timestamps is formulated as $X=\{X_{1},X_{2},…,X_{n},…,X_{N}\} \in \mathbb{R}^{N{\times} T{\times}C}$,where $C$ is the number of feature channels including the target variable $Patv$. For the $n$-th turbine, we denote $X_{n}=\{X_{n}^{1},X_{n}^{2},…,X_{n}^{T}\} \in \mathbb{R}^{T{\times}C}$ as the context records of all timestamps for the $n$-th turbine. Symmetrically, the context records of the whole farm at timestamp $t$ is denoted as $X^{t}=\{X_{1}^{t},X_{2}^{t},…,X_{N}^{t}\} \in \mathbb{R}^{N{\times}C}$.
\begin{definition}{\rm Wind Power Forecasting.}
Assuming that the current timestamp is $t$, the wind power forecasting problem is to predict all turbines' power generations of future $F$ timestamps utilizing the historical dynamic context factors of previous $H$ timestamps, which is also a spatio-temporal data prediction problem. Mathematically, the predictions $\hat{Y}_{t+F}=\{Y_{1}^{1},Y_{1}^{2},…,Y_{2}^{1},Y_{2}^{2},…,Y_{N}^{F}\} \in \mathbb{R}^{N{\times}F{\times}1}$, where $1$ represents the target variable $Patv$, is obtained:
\begin{equation}
\hat{Y}_{t+F}=f(X_{t-H} \mid \Phi)
\end{equation}
where $f(\cdot)$ represents the model for wind power forecasting, $X_{t-H}$ denotes the historical dynamic context records and $\Phi$ represents the parameters in our model.
\end{definition}

\section{Method}
The overall architecture of the proposed framework HSTTN is shown in Figure~\ref{fig:overview}, which is composed of the hourglass-shaped encoder-decoder architecture, the residual spatiotemporal encoder/decoder layer, the Contextual Fusion Block and a wind power regression module. The hierarchical residual spatiotemporal encoder/decoder layers (RSTEL/RSTDL) with pooling and up-convolution operations capture multiscale temporal dependencies and global spatial correlations from the embedded temporal-wise features and spatial-wise features respectively. Skip-connections between encoder and decoder will help enhance finer predictions by recovering localized coarse temporal information.
Meanwhile, the CFBs inserted in residual spatiotemporal layers take the outputs of each temporal sublayer and spatial sublayer to capture the contextual information interaction between spatial and temporal features and propagate the enhanced representations carrying both spatial and temporal information. The encoder-decoder modeling paradigm generates output sequence one element at a time.

\subsection{Hourglass-shaped Encoder-decoder} \label{subsec:subsec_4_2}
This is the main body of our framework. To tackle the long-term time series forecasting problem, global self-attention mechanisms are always preferred for modeling dependencies without regard to their distance in the sequences. Different from previous local modeling works~\cite{yu2020probabilistic,ZhuChen}, we try to capture the global spatial correlations among different locations, as wind characteristics are similar in a local area or distant areas with similar climatic conditions. Besides, turbines that are not close to each other but sharing the same working status will also perform in an analogous way. So we employ the transformer architecture to model both long-range temporal dependencies and spatial correlations. Noted that the inputs of the first decoder layer $X^{d}_{t+H}$ are different from encoder inputs $X^{e}_{t-H}$, which contain only future time spots and turbine locations without knowing the meteorological data and turbine internal status. The unknown factors in decoder's inputs are padded with zero.

\begin{figure}[t]
    \centering
    \includegraphics[width=0.90\linewidth]{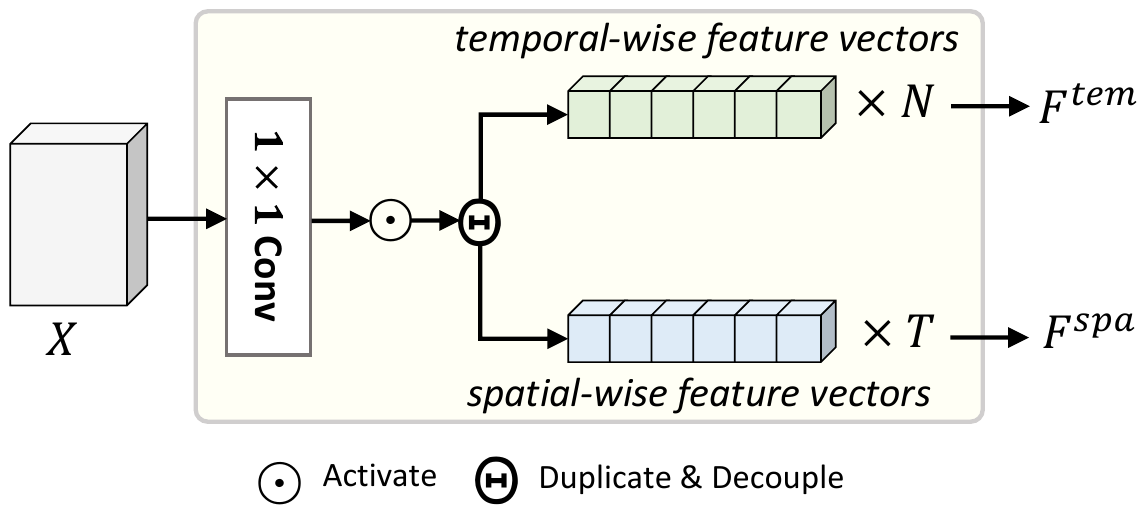}
    \vspace{-3mm}
    \caption{The details of raw feature embedding module.}
    \vspace{0mm}
    \label{fig:embedding}
\end{figure}

As introduced in Section~\ref{sec:problem_specification}, the raw wind power context records are spatiotemporal structured data. As is depicted in Figure~\ref{fig:embedding}, a $1{\times}1$ convolutional block is firstly adopted to learn a high-dimension latent feature embedding from raw inputs. Without padding and stride, this step will generate a 2-D feature map $F^{Conv} \in \mathbb{R}^{N{\times}T{\times}d_{model}}$, where $d_{model}$ is the number of convolution kernels, i.e., the embedded features dimension:
\begin{equation}
    {F^{Conv}}^{i}=ReLU(W^{i} \star X + b^{i}),i=1,2,…,d_{model},
\end{equation}
where ${F^{Conv}}^{i}$ denotes the $i$-th channel in the final feature map $F^{Conv}$, $ReLU(\cdot)$ is the activation function, $\star$ represents the convolution operation, $W^{i} \in \mathbb{R}^{1{\times}1{\times}C}$ and $b^{i} \in \mathbb{R}^{C}$ are the learned weights of the $i$-th kernel and bias, respectively. Then $F^{Conv}$ is duplicated and decoupled into $N$ temporal-wise feature vectors $F^{tem} \in \mathbb{R}^{T{\times}d_{model}}$ and $T$ spatial-wise feature vectors $F^{spa} \in \mathbb{R}^{N{\times}d_{model}}$ as the inputs of residual encoder layers.

\begin{figure}[t]
    \centering
    \includegraphics[width=1.0\linewidth]{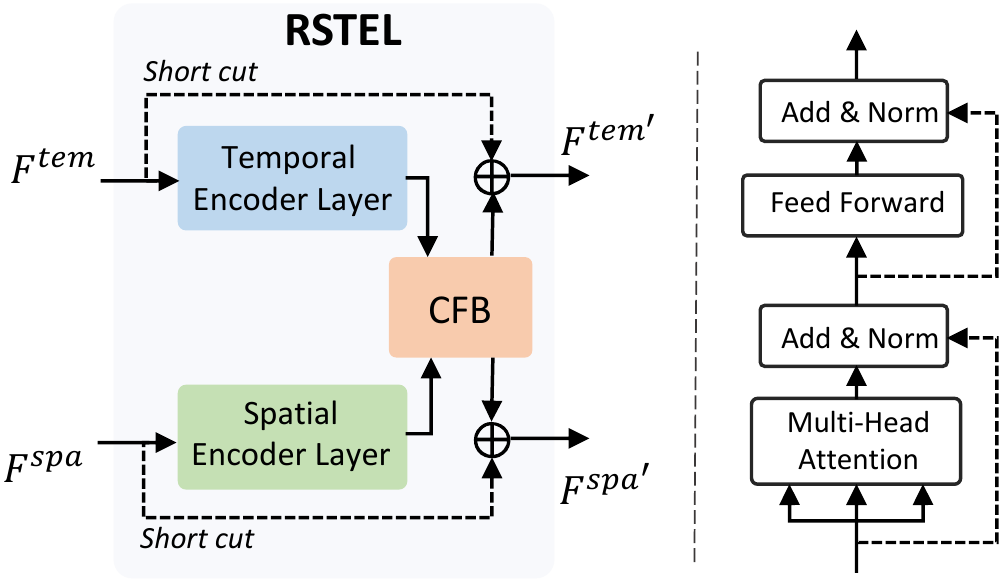}
    \vspace{-7mm}
    \caption{The structure of Residual Spatiotemporal Encoder Layer (RSTEL). $\bigoplus$ represents element-wise addition.}
    \vspace{0mm}
    \label{fig:RSTEL}
\end{figure}

Significantly, different from the general transformer architecture, we propose to successively downsample the long-term fine-grained temporal scale to coarse-grained scale in the encoder and recovering it with upsamplings in the decoder, which forms a hierarchical structure with different temporal scales. The cascaded residual spatiotemporal encoder/decoder layers which can self-repeat multiple times form the hourglass-shaped network and the details of RSTEL is depicted in Figure~\ref{fig:RSTEL}. The input features $F^{tem}$ and $F^{spa}$ are firstly fed to the corresponding temporal encoder layer and spatial encoder layer, which apply the standard Multi-head Self-Attention (MSA) mechanism~\cite{VaswaniSha}. Taking temporal encoder layer for example, the $i$-th temporal encoder layer's input $F_{i}^{tem}$ is first fed into three linear layers to generate query, key and value embedding: $Q_{i}^{tem}=F_{i}^{tem} W_{i}^{q}$, $K_{i}^{tem}=F_{i}^{tem} W_{i}^{k}$, $V_{i}^{tem}=F_{i}^{tem} W_{i}^{v}$, where $W_{i}^{q} \in \mathbb{R}^{d_{model}{\times}d_{k}}$, $W_{i}^{k} \in \mathbb{R}^{d_{model}{\times}d_{k}}$, and $W_{i}^{v} \in \mathbb{R}^{d_{model}{\times}d_{v}}$ are the projection parameter matrices. we can obtain the latent representation:
\begin{equation}
\begin{split}
    head_{i}^{tem}&=Softmax(\cfrac{Q^{tem}{K^{tem}}^{T}}{\sqrt{d_{k}}})V^{tem}, \\
    Attn^{tem}&=Concat(head_{1}^{tem},…,head_{h}^{tem})W^{O},
\end{split}
\end{equation}
where $Attn^{tem} \in \mathbb{R}^{T{\times}d_{model}}$. To be exact, for each embedded $F^{Conv}$, there are $N$ encoded vectors $Attn^{tem}$. Then, both temporal and spatial features are delivered into the contextual fusion block, which generates the enhanced representation. At last the origin inputs are directly added to the fusion output in a residual manner, which can help to reduce overfitting and gradient vanishing. Therefore the output of the $i$-th residual spatiotemporal encoder layer can be written as follows:
\begin{equation}
\begin{split}
    {F^{tem}_{i}}^{'}&=FUSE(Attn^{tem}_{i}) + F^{tem}_{i},\\
    {F^{spa}_{i}}^{'}&=FUSE(Attn^{spa}_{i}) + F^{spa}_{i},
\end{split}
\end{equation}
where $FUSE(\cdot)$ represents the contextual fusion in Section~\ref{subsec:subsec_4_3}. After the maxpooling, we have the inputs for the next RSTEL $F^{tem}_{i+1} \in \mathbb{R}^{\frac{T}{p}{\times}d_{model}}$, where $p$ is the pooling factor.
RSTDL follows a similar structure except for an additional multi-head attention over the outputs of its corresponding RSTEL. The inter-scale skip connections between encoder and decoder layers further facilitate the model to make precise predictions. Specifically, the outputs of each RSTEL are directly concatenated to the outputs of their corresponding RSTDL, followed by a up convolution to recover the fine-grained details.

\begin{figure}[t]
    \centering
    \includegraphics[width=1.0\linewidth]{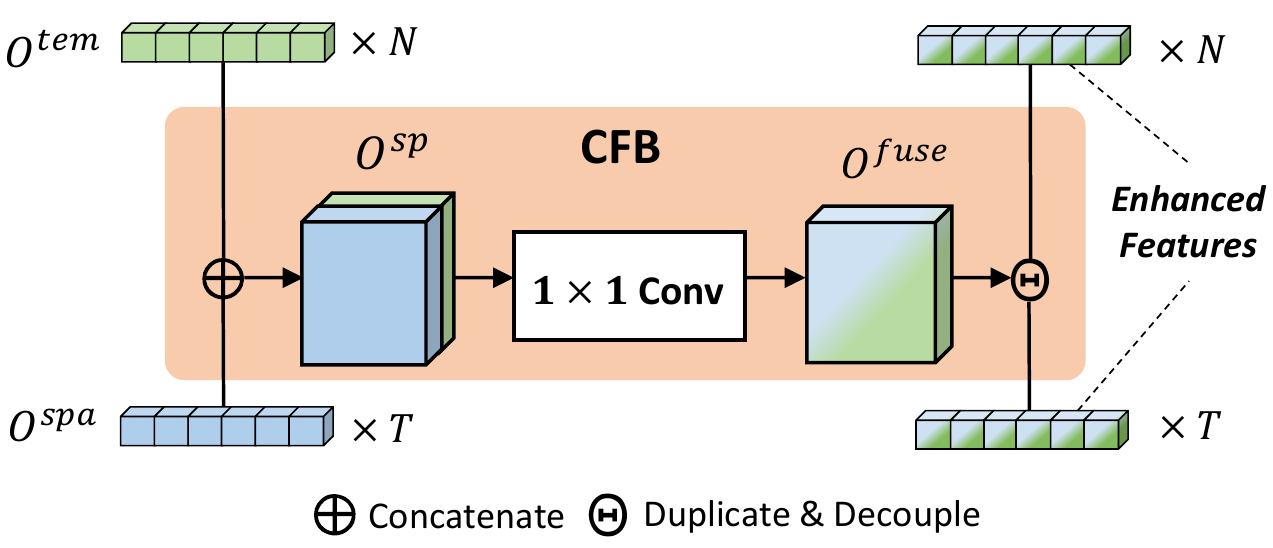}
    \vspace{-7mm}
    \caption{The structure of our Contextual Fusion Block (CFB).}
    \label{fig:cfb_block}
    \vspace{0mm}
\end{figure}

\subsection{Contextual Fusion Block} \label{subsec:subsec_4_3}
The temporal dependencies and spatial correlations are characterized in their own branches thus lack of contextual information interactions.
Here we utilize a simple, computationally efficient but effective module named Contextual Fusion Block (CFB) to capture the contextual joint features and propagate them among temporal and spatial representations. Figure~\ref{fig:cfb_block} shows the structure of CFB.

We take the feature fusion in encoder for example since the decoder shares the same mechanism. As described in Section~\ref{subsec:subsec_4_2}, the original input features are decoupled into multiple sequences of vectors. After aggregating information from the whole sequence, the outputs of each temporal encoder layer are ${Attn^{tem}}^{n} \in \mathbb{R}^{T{\times}d_{model}},n=1,2,…,N$ and the outputs of the corresponding spatial encoder layer are ${Attn^{spa}}^{t} \in \mathbb{R}^{N{\times}d_{model}},t=1,2,…,T$. Both of them are firstly stacked to the original 2-D shape feature maps and then concatenated along the channels: 
\begin{equation}
\begin{split}
O^{spa}&=Concat({Attn^{spa}}^{1},…,{Attn^{spa}}^{T}), \\
O^{tem}&=Concat({Attn^{tem}}^{1},…,{Attn^{tem}}^{N}), \\
O^{sp}&=Concat(O^{spa^{T}},O^{tem}),
\end{split}
\end{equation}
where $Concat(\cdot)$ denotes a concatenation operation, $O^{spa} \in \mathbb{R}^{T{\times}N{\times}d_{model}}$ and $O^{tem} \in \mathbb{R}^{N{\times}T{\times}d_{model}}$ are the stacked spatial encoder layer output and temporal encoder layer output respectively and $O^{sp} \in \mathbb{R}^{N{\times}T{\times}d_{model}{\cdot}2}$ is the learned representation including both spatial and temporal characteristics. Then, in order to remove redundant information and reduce feature dimensions, a $1{\times}1$ convolution is employed to capture the contextual spatio-temporal correlations and generate the enhanced representation $O^{fuse} \in \mathbb{R}^{N{\times}T{\times}d_{model}}$:
\begin{equation}
    {O^{fuse}}^{i}=ReLU(W^{i} \star O^{sp} + b^{i}),i=1,2,…,d_{model},
\end{equation}
where ${O^{fuse}}^{i}$ is the $i$-th channel in $O^{fuse}$. Finally, the enhanced informative features carrying both spatial and temporal information are duplicated and decoupled into multiple temporal-wise and spatial-wise sequence vectors as described in Section~\ref{subsec:subsec_4_2}, which are then fed to their corresponding next encoder layers for higher-level representation learning.

\subsection{Wind Power Regression Module}
The outputs of the original scale residual spatiotemporal encoder and decoder layers are concatenated to make the final predictions. The concatenated outputs $O^{orign} \in \mathbb{R}^{N{\times}F{\times}d_{model}{\cdot}2}$, are fed into a fully-connected layer to predict wind power generations of the next $F$ timestamps ${\hat{Y}}_{t+F} \in \mathbb{R}^{N{\times}F{\times}1}$ :
\begin{equation}
    {\hat{Y}}_{t+F}=Drop(O^{orign})W^{Y}+b^{Y},
\end{equation}
where $Drop$ is the dropout operation, $W^{Y} \in \mathbb{R}^{2{\cdot}d_{model}{\times}1}$ are the learnable parameters and  $b^{Y}$ is the bias.

To evaluate the difference between our prediction and the ground truth, we utilize mean square error (MSE) as our loss function, which is defined below:
\begin{equation}
    Loss=MSE=\frac{1}{m}\sum_{i=1}^{m}(Y_{t+F}^{i} -{\hat{Y}}_{t+F}^{i})^{2},
\end{equation}
where $m$ is the number of samples.

\section{Experiments}\label{sec:experiments}
\subsection{Experimental Setup}
\textbf{Datasets:} We conduct experiments on two challenging real-world datasets:
i) \textbf{SDWPF} \cite{zhou2022sdwpf} is obtained from the real-world data from Longyuan Power Group Corp. Ltd. This dataset contains 4,727,520 records sampled every 10 minutes and are collected from a wind farm with 134 wind turbines in 245 days. Each record contains 13 attributes including critical external features (such as wind speed, wind direction and temperature) and essential internal features (such as inside temperature, nacelle direction and so on).
ii) \textbf{Engie}\footnote{https://opendata-renewables.engie.com} is obtained from ENGIE group. The wind power data consists of 1,057,968 records from 1 January 2013 to 12 January 2018, obtained by sampling every 10 minutes from a wind farm containing 4 wind turbines. Each record contains 34 attributes.

\noindent\textbf{Implementation Details:}
In our experiment, we utilize the historical records of 144 time slots to forecast the wind power generations in the next 144 time slots. For SDWPF, we sequentially split the dataset into 155 days, 30 days and 60 days for training, validation, and testing, respectively. For Engie, the dataset is split into 1296 days, 180 days and 360 days for training, validation and testing, respectively. Finally, the whole dataset is normalized with Z-score Normalization and inverted to the original scale when performing evaluation.
Our proposed model is implemented with the PyTorch framework. The feature dimension $d_{model}$ for multi-head attention is set to 16 and the number of head is set to 2 for both datasets. The number of convolution kernels in CFB is also 16. For this 144 time slot prediction task, we downsampling the original scale with 3 and 2 times successively in encoder, while upsampling it to the original scale in decoder. The initial learning rate is 1e-4 and 1e-3 and decreases gradually. Adam optimizer is adopted to minimize the MSE loss until the training process is ended by early stopping strategy. We choose the best model on validation set as the final model and evaluate its performance on testing set for fair comparisons.

\noindent\textbf{Evaluation Metrics:}
We use two widely used metrics including mean average error (MAE) and root mean square error (RMSE) to evaluate all methods' performance on the whole wind farm. They are defined as:
{\small
\begin{equation} \label{eq:metrics}
\begin{split}
    MAE&=\frac{1}{m}\sum_{n=1}^{N}\sum_{i=1}^{m}\lvert y^{(n,i)}-\hat{y}^{(n,i)} \rvert,\\
    RMSE&=\sum_{n=1}^{N}\sqrt{\frac{1}{m}\sum_{i=1}^{m}{(y^{(n,i)}-\hat{y}^{(n,i)})}^2},
\end{split}
\end{equation}
}
where $N$ is the number of turbines and $m$ is the number of samples of each turbine.
Besides, SDWPF introduced several invalid conditions of the records caused by its recording system. These invalid values will not be used in evaluation.

\subsection{Baselines}
In this paper, we compare the proposed HSTTN against seven deep learning methods, including two RNN-based models LSTM~\cite{hochreiter1997long}, GRU~\cite{gru}, two Transformer-based models: Informer~\cite{ZhouZhang}, TSAT~\cite{ng2022expressing}, and three spatial-temporal forecasting models: Baidu* \footnote{https://github.com/PaddlePaddle/PGL/tree/main/examples/\\kddcup2022/wpf\_baseline}, DCRNN~\cite{li2018dcrnn_traffic}, Bi-STAT~\cite{chen2022bidirectional}. All methods are implemented on both preprocessed datasets under the same experimental setup for a fair comparison.

\begin{table}
\renewcommand\arraystretch{1.25}
\centering
\caption{Performance of different methods on both datasets.}
\vspace{-3mm}
\resizebox{1.0\columnwidth}{!}{
\begin{tabular}{c|cc|cc}
\hline
\multirow{2}{*}{Method} & \multicolumn{2}{c|}{SDWPF} & \multicolumn{2}{c}{Engie}  \\
\cline{2-5}
& MAE(MW) & RMSE(MW) & MAE(MW) & RMSE(MW) \\
\hline
\hline
LSTM&41.23&46.15&1.34&1.58  \\
GRU	&40.92	&46.40	&1.35	&1.56 \\
Baidu&	37.73	&43.72	&1.16	&1.36\\
DCRNN	&38.33&	46.52&	1.05	&1.22\\
Informer	&37.15	&43.45	&1.15	&1.30\\
Bi-STAT	&38.34	&45.75	&1.03&	1.23\\
TSAT	&38.03	&44.91	&1.10	&1.28\\
\hline
\textbf{HSTTN}	&\textbf{33.07}	&\textbf{40.16}	&\textbf{0.91}	&\textbf{1.08} \\
\hline
\end{tabular}}
\vspace{0mm}
\label{tab:all_methods}
\end{table}

Table~\ref{tab:all_methods} summarizes the performance of all comparison methods on SDWPF and Engie respectively, and our proposed HSTTN performs the best on both datasets. LSTM, GRU, Informer and TSAT are time series forecasting models which lack spatial feature modeling. So we decouple the spatiotemporal input data to multiple temporal sequences in the manner mentioned in Section~\ref{subsec:subsec_4_2} and feed them into these models. As a result, our HSTTN achieves the lowest MAE and RMSE on both dataset.  We can observe that when handling WPF as multivariate time series forecasting, Transformer-based models (Informer, TSAT, Baidu*, Bi-STAT and HSTTN) outperforms those RNN-based models LSTM, GRU and which imply the  effectiveness of self-attention for capturing long-range temporal dependencies. Informer and TSAT also utilize self-attention to learn temporal representations but lack spatial information modeling, which limits their performance. Compared with simple RNN, DCRNN improves the performance to some degree with the help of learning spatial representations explicitly, which demonstrates the importance of spatial features for multi-turbine wind power forecasting. By modeling both spatial-temporal context, DCRNN and Bi-STAT outperforms common recurrent neural networks (LSTM, GRU) and temporal transformer models (Informer, TSAT) and reaches comparable results with our HSTTN on Engie dataset, but the performance is not satisfactory on SDWPF dataset. The reason is that DCRNN and Bi-STAT capture spatial context based on Euclidean connectivity and distance, while the turbine distributions on SDWPF are much more complex than that of Engie and the spatial context of wind power is also related to meteorological conditions and turbine status. Compared to the above methods, we not only capture the global and comprehensive spatial-temporal correlations by self-attention mechanism but also carefully designed the hourglass-shaped network architecture for long-term prediction, thereby achieving the state-of-the-art performance for both dataset.

\begin{table}
\renewcommand\arraystretch{1.25}
\centering
\caption{Performance of different temporal scales.}
\vspace{-3mm}
\resizebox{1.0\columnwidth}{!}{
\begin{tabular}{c|cc|cc}
\hline
\multirow{2}{*}{Variant} & \multicolumn{2}{c|}{SDWPF} & \multicolumn{2}{c}{Engie}  \\
\cline{2-5}
& MAE(MW) & RMSE(MW) & MAE(MW) & RMSE(MW) \\
\hline
\hline
STTN &35.29  &41.57  &1.03  &1.23  \\
2-STTN  &34.83  &41.22  &0.98  &1.20 \\
4-STTN &36.18  &42.29  &1.07  &1.27  \\
NoSkip &35.98  &41.86  &1.04  &1.25  \\
\hline
\textbf{HSTTN} & \textbf{33.07} & \textbf{40.16} & \textbf{0.91} & \textbf{1.08} \\
\hline
\end{tabular}}
\vspace{0mm}
\label{tab:ablation_ms}
\end{table}

\subsection{Ablation Studies}
In this subsection, we perform extensive analyses to verify the effectiveness of each component of the proposed HSTTN.
\subsubsection{Effectiveness of Hierarchical Temporal Learning}
\begin{itemize}
    \item Spatial-Temporal Transformer Network (STTN): In this variant, we remove the downsampling, upsampling and skip-connection operations to explore the performance of the original temporal scale only.
    \item Two scale Spatial-Temporal Transformer Network (2-STTN): This variant implement the downsampling and upsampling once each with the factor of 3, which leading to 2 temporal scale learning, to explore the effectiveness of coarse-grained semantic representations.
    \item Four scale Spatial-Temporal Transformer Network (4-STTN): Similarly, we implement the downsampling the upsampling 3 times with the factors of 3, 2 and 2.
    \item HSTTN without skip-connections (NoSkip): To demonstrate the effectiveness of the skip-connections between encoder and decoder, we remove them in this variant.
\end{itemize}
In Table~\ref{tab:ablation_ms}, we can observe that 2-STTN and HSTTN both perform better than STTN, which demonstrates the effectiveness of both the coarse-grained temporal dependencies and our hierarchical architecture. But the 4-STTN variant can't make further improvement, which may suffer from overfitting problems as the network goes deeper. The HSTNN outperforms NoSkip variant, which proves that aggregating information from different scales enables the model to make more precise predictions.

\begin{table}
\renewcommand\arraystretch{1.25}
\centering
\caption{Performance of different variants of HSTTN and variants of CNN model.}
\vspace{-3mm}
\resizebox{1.0\columnwidth}{!}{
\begin{tabular}{c|cc|cc}
\hline
\multirow{2}{*}{Variant} & \multicolumn{2}{c|}{SDWPF} & \multicolumn{2}{c}{Engie}  \\
\cline{2-5}
& MAE(MW) & RMSE(MW) & MAE(MW) & RMSE(MW) \\
\hline
\hline
T-CNN  &39.25  &44.83  &1.20  &1.38  \\
S-CNN  &41.21  &46.02  &1.28  &1.46 \\
T-Only &35.50  &41.78  &1.03  &1.25  \\
S-Only  &40.01  &45.63  &1.25  &1.44 \\
ST-Only &35.21  &41.60  &1.02  &1.25 \\
\hline
\textbf{HSTTN} & \textbf{33.07} & \textbf{40.16} & \textbf{0.91} & \textbf{1.08} \\
\hline
\end{tabular}}
\vspace{0mm}
\label{tab:ablation}
\end{table}

\subsubsection{Importance of Capturing Global Information}
\begin{itemize}
    \item Temporal 1-D CNN Only (T-CNN): To verify the importance of global temporal information, we replace the transformer in the above T-Only with stacked 1-D convolutional neural networks to capture local temporal dependencies then make predictions.
    \item Spatial 1-D CNN Only (S-CNN): We replace the transformer in the above S-Only with stacked 1-D CNN to verify the significance of global spatial correlations.
    \item Temporal Transformer Only (T-Only): This variant only includes temporal Transformer layers to verify the effectiveness of our global temporal feature modeling and hierarchical temporal learning.
    \item Spatial Transformer Only (S-Only): We implement this variant including only spatial transformer layers to explore the effectiveness of global spatial information.
\end{itemize}
The performance of different variants is shown in Table~\ref{tab:ablation}. We can observe that T-Only can achieve a comparable result that contribute most to our HSTTN, which demonstrates the effectiveness of our hourglass-shaped hierarchical framework for capturing long-range temporal dependencies. We explore to extract spatial temporal features by CNN structure, which is widely used for local feature extraction. The performance of T-CNN and S-CNN can not match with T-Only ,S-Only and not to mention HSTTN on both datasets, which proves the importance of global information for wind power forecasting. Spatial features is only supplemental to WPF, so S-Only and S-CNN performs poorly.

\subsubsection{Effectiveness of Spatial-Temporal Contextual Fusion}
\begin{itemize}
    \item Spatial-Temporal Transformer Only (ST-Only): To verify the effectiveness of our CFB, we implement a simple fusion variant without the contextual fusion block integrated in between. Then the outputs of the last RSTDL are simply concatenated to make predictions.
\end{itemize}
The experiments results are illustrated in Table \ref{tab:ablation}. ST-Only slightly improves T-Only demonstrates the effectiveness of the global spatial information. Our HSTTN outperforms both T-Only and ST-Only reveals that the temporal and spatial features can not be casually combined and our CFBs that properly fuses spatiotemporal context features is effective.

\begin{table}
\renewcommand\arraystretch{1.25}
\setlength{\tabcolsep}{1mm}
\centering
\caption{Performance of three primary hyperparameters.}
\vspace{-3mm}
\resizebox{1.0\columnwidth}{!}{
\begin{tabular}{c|c|cc|cc}
\hline
{Hyper-} & \multirow{2}{*}{Settings} & \multicolumn{2}{c|}{SDWPF} & \multicolumn{2}{c}{Engie}  \\
\cline{3-6}
 parameter & & MAE(MW) & RMSE(MW) & MAE(MW) & RMSE(MW)  \\
\hline
\hline
\multirow{3}{*}{Kernel Size} & 1×1  &\textbf{33.07}  &\textbf{40.16}  & \textbf{0.91} & \textbf{1.08}  \\
                             & 3×3  &35.94  &42.65  & 0.96 & 1.20  \\
                             & 5×5  &36.29  &43.04  & 1.10 & 1.30  \\
\hline
\multirow{3}{*}{Layer Num}   & 1, 1 &33.67  &40.45  & 0.96  & 1.12  \\
                             & 2, 1 &\textbf{33.07}  &\textbf{40.16}  & 0.95  & 1.10  \\
                             & 2, 2 &34.92  &41.03  & \textbf{0.91} & \textbf{1.08}  \\
\hline
\multirow{4}{*}{Dimensions}  & 8   & 36.17  & 42.16  & 0.95 & 1.12  \\
                             & 16  &\textbf{33.07}  &\textbf{40.16}  & \textbf{0.91} & \textbf{1.08}  \\
                             & 32  & 35.41  & 41.64  & 1.03 & 1.23  \\
                             & 64  & 38.67  & 44.47  & 1.12 & 1.32  \\
\hline
\end{tabular}}
\vspace{0mm}
\label{tab:hyper}
\end{table}

\subsubsection{Analysis of different hyperparameters}
We conduct extensive experiments on three important hyperparameters in HSTTN to find the best settings.
\begin{itemize}
    \item Kernel Size: the kernel size in contextual fusion block.
    \item Layer Num: the number of repeated residual spatiotemporal encoder/decoder layers.
    \item Dimensions: the number of embedded feature dimensions after the $1{\times}1$ convolution.
\end{itemize}
Table~\ref{tab:hyper} records the hyperparameters settings and results. According to the results, we decide the kernel size, encoder layers, decoders layers and embed feature dimension as 1, 2, 1 and 16 for SDWPF and as 1, 2, 2, 16 for Engie.

\section{Conclusion}\label{sec:conclusion}
In this work, we propose a hourglass-shaped encoder-decoder model termed Hierarchical Spatial-Temporal Transformer Network to deal with the challenging long-term wind power forecasting problem. The model design is motivated by two main limitations of existing works. First, most of these works are designed for short-term while lack of effective long-term prediction solutions. Second, existing wind power forecasting works lack properly designed module for spatial-temporal context feature mining. Thus, we adopt Transformer mechanism to capture both long-range temporal dependencies and global spatial correlations and carefully design a hierarchical temporal learning structure to facilitate long-term forecasting with complementary coarse-grained semantics. Moreover, we design a Contextual Fusion Block to further enhance the learned features and improve the performance.

\section*{Acknowledgments}
This work was supported in part by the Guangdong Basic and Applied Basic Research Foundation (NO. 2020B1515020048), in part by the National Natural Science Foundation of China (NO. 61976250), in part by the Shenzhen Science and Technology Program (NO. JCYJ20220530141211024) and in part by the Fundamental Research Funds for the Central Universities under Grant 22lgqb25.

\bibliographystyle{named}
\bibliography{ijcai23}

\end{document}